\documentclass[letterpaper, 10 pt, conference]{ieeeconf}  

\IEEEoverridecommandlockouts                              

\usepackage{graphics} 
\usepackage{multirow}
\graphicspath{ {imagenes/}  }
\usepackage{subcaption} 
\usepackage{epsfig} 
\usepackage{amsmath} 
\usepackage{amssymb}  
\usepackage{hyperref}
\hypersetup{
    colorlinks=true,
    linkcolor=black,
    urlcolor=blue,
    }

\usepackage[ruled]{algorithm2e}
\usepackage{algorithmicx}
\usepackage{algpseudocode}

\DeclareMathAlphabet{\mathpzc}{OT1}{pzc}{m}{it}

\usepackage[table,xcdraw]{xcolor}
\usepackage{color,soul}

\usepackage{setspace}

\title{\LARGE \bf Categorized Grid and Unknown Space Causes for\\ LiDAR-based Dynamic Occupancy Grids}

\author{Víctor Jiménez-Bermejo$^{1}$, Jorge Godoy$^{1}$, Antonio Artuñedo$^{1}$, and Jorge Villagra$^{1}$
\thanks{*This work has been funded by MCIN/AEI/10.13039/501100011033 and by ERDF A way of making Europe with the project DISCERN (PID2021-125850OB-I00).}
\thanks{$^{1}$The authors are with the Centro de Automática y Robótica, CSIC-Universidad Politécnica de Madrid, 28500 Arganda del Rey, Madrid, Spain. (email: victor.jimenez@csic.es; jorge.godoy@csic.es; antonio.artunedo@csic.es; jorge.villagra@csic.es).
}
}

\begin{document}

\maketitle
\thispagestyle{empty}
\pagestyle{empty}

\begin{abstract}
Occupancy Grids have been widely used for perception of the environment as they allow to model the obstacles in the scene, as well as free and unknown space. 
Recently, there has been a growing interest in the unknown space due to the necessity of better understanding the situation.
Although Occupancy Grids have received numerous extensions over the years to address emerging needs, currently, few works go beyond the delimitation of the unknown space area and seek to incorporate additional information. 
This work builds upon the already well-established LiDAR-based Dynamic Occupancy Grid to introduce a complementary Categorized Grid that conveys its estimation using semantic labels while adding new insights into the possible causes of unknown space. The proposed categorization first divides the space by occupancy and then further categorizes the occupied and unknown space. Occupied space is labeled based on its dynamic state and reliability, while the unknown space is labeled according to its possible causes, whether they stem from the perception system's inherent constraints, limitations induced by the environment, or other causes. The proposed Categorized Grid is showcased in real-world scenarios demonstrating its usefulness for better situation understanding.
\end{abstract}

\section{Introduction}
\label{sec:introduction}

Perception of the environment is a challenging task essential for autonomous driving. 
Within the fields of obstacle detection and navigable space estimation, one of the most popular strategies is the calculation of Occupancy Grids (OG) \cite{1989_elfes} using LiDAR data. These enable accurate modeling of every obstacle in the scene while also delimitating the free and unknown space. 

OG-based approaches have continuously evolved to accommodate new emerging requirements. For example, in autonomous driving applications, perhaps the most notable advance is the estimation of obstacles' dynamics---Dynamic Occupancy Grids (DOG)---as it allows properly addressing the common dynamic driving scenarios \cite{2017_Nuss, 2018_Steyer_grid, 2022_Schreiber}. Many other works incorporate additional information about the occupied space, such as obstacle classification \cite{2022_Schreiber, 2023_victor_framework, 2019_Erkent}, individual identification over time \cite{2020_Danescu}, or modeling of future coverage areas \cite{2021_Juan}.
Similarly, several studies also feature new data on the free space, especially regarding the driveable zone, including free roads and sidewalk \cite{2019_Erkent, 2019_Lu, 2015_Kurdej}.

In contrast, fewer works address the inclusion of additional information about the unknown space. \cite{2015_Kurdej} and \cite{2017_Hoermann} take advantage of digital maps to fill unknown areas with prior information regarding the road layout or permanent obstacles, such as buildings. Schreiber et al. \cite{2022_Schreiber} predict the drivable area within the unknown space using recurrent neural networks. \cite{2018_Steyer_grid} derives a related concept in which the measured free space is predicted in the next iteration as ``passable area'' due to its time-dependent condition in dynamic scenarios.

Currently, information about unknown space is gaining special interest in order to achieve a better understanding of the situation. 
Therefore, despite not being explicitly grounded in OGs, there are some other works that infer the unknown space and provide new insights. 
\cite{2018_Orzechowski} computes the sensing field of the ego-vehicle using the field of view (FoV) of its sensors and considering occlusions caused by obstacles. It intersects this sensing field with a road map to identify occluded areas within the road lanes. Additionally, it labels these areas as either relevant or irrelevant based on the possible interactions with the ego-vehicle. Similarly, \cite{2021_Koschi} also computes these borders, but denotes the potential obstacles that can appear through them and are relevant for the motion of the ego-vehicle. 
Sanchez et al. \cite{2020_Sanchez, 2022_tesis_Sanchez} reduce the gap between perception and decision-making modules by transforming the outcome of the perception system into an occupancy-based lane grid that also categorizes the unknown space based on different notions. First, the cause is distinguished between occlusions and FoV constraints. Then, based on interactions with other vehicles, it can also be categorized as ``safety area'' (the space in front of a moving vehicle must be obstacle-free for reglamentary safety reasons) or ``protected area'' (space with circulation obstructed due to the presence of a detected vehicle). 

\begin{figure*}[!htb]
    \centering
    \includegraphics[width=\textwidth]{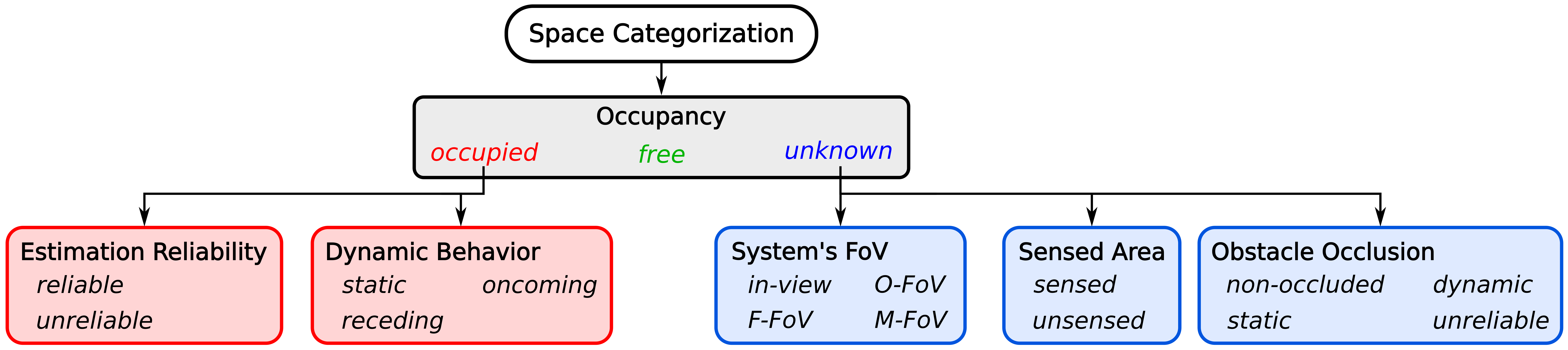}    
    \caption{Scheme of the proposed space categorization. M-FoV/O-FoV/F-FoV: Maximum, Occupied, and Free Field of View.}
    \label{fig:categorization}
\end{figure*}

As discussed previously, despite the increasing interest, there is limited exploration into incorporating additional information about the unknown space, especially within the field of OGs.  
Furthermore, most existing works focus on identifying what may exist within the unknown space or its implications, rather than explaining the rationale behind the system's estimation. 
Therefore, this work proposes 
(i)~an in-depth analyses that elucidate the possible causes of unknown space within the already well-established DOGs and 
(ii)~a complementary Categorized Grid (CG) that conveys the estimation of the DOG and the new unknown space information in an easily understandable manner.

The remainder of this paper is structured as follows. Section~\ref{sec:DOG} briefly introduces the basics of the DOG used in this work as input data. Then, in Section~\ref{sec:CG} the proposed CG is described in detail. Section~\ref{sec:experiments} presents the experiments conducted for validation. Lastly, Section~\ref{sec:conclusions} summarizes the work presented and provides an outlook for future work.

\section{Dynamic Occupancy Grid Background}
\label{sec:DOG}

The DOG employed in this work is based on the approaches presented in \cite{2017_Nuss, 2023_victor_framework}. It divides the environment surrounding the ego-vehicle into small square cells that store occupancy and dynamic states. 

The occupancy estimation is conducted using the Dempster Shafer theory, being the frame of discernment the events of occupied and free ($\Omega = \{O, F\}$). Thus, the occupancy state of each cell $c$ contains a mass for occupied $m(O^c)$, a mass for free $m(F^c)$, and a mass for unknown $m(\Omega^c) = 1 - m(O^c) - m(F^c)$. In each iteration, the last captured LiDAR scans are modeled in terms of occupancy and used to update the occupancy state of the cells. 

The dynamic state estimation of the obstacles in the scene is addressed using a particle filter. Particles ($\rho$) can move freely along the grid, but their weights are directly related to the cells' occupied mass. Therefore, particles' movements model the prediction of occupied mass displacement, while the updated occupied mass is used to adjust particles' weights. 
The dynamic state of a cell $\pmb{v}^c$ is calculated based on the set of particles within it. Cells can be divided into static and dynamic using the Mahalanobis criterion, see \cite{2023_tesis_Victor}. 

Additionally, in order to obtain a reference of the obstacles' height, the elevation of LiDAR points is also rasterized into the grid cells. Thus, each cell also stores the height of its highest and lowest points.

For further details on the DOG basis, the reader may refer to the works cited above.

\section{Categorized Grid}
\label{sec:CG}

Categorizing or labeling the space is a useful tool that many works have employed directly or indirectly to facilitate the understanding and management of OGs' data. For example, cells are commonly divided by occupancy state into \textit{occupied}, \textit{free} or \textit{unknown} and by dynamic state (in DOGs) into \textit{static} or \textit{dynamic} \cite{2023_victor_framework, 2017_Piewak, 2022_Banfi}. 

The proposed CG consists of a multi-labeled grid based on the space categorization depicted in Fig.~\ref{fig:categorization}. First, the estimated environment is divided by occupancy state. Then, \textit{occupied} and \textit{unknown} cells are further categorized with respect to different features. \textit{Occupied} cells are evaluated regarding their dynamic state and estimation reliability; while \textit{unknown} cells are analyzed with respect to their possible causes. 

Labels within the same block are mutually exclusive, for example, a cell is either \textit{occupied}, \textit{free}, or \textit{unknown}; or is either \textit{reliable} or \textit{unreliable}. In contrast, labels that share occupancy status but are in different feature blocks may co-exist, e.g. a cell can be \textit{occupied}, \textit{reliable} and \textit{static}; or \textit{unknown}, \textit{in-view}, \textit{unsensed} and \textit{occluded-static}.

Thus, the CG ($\mathcal{G}_\psi$) is a grid in which each cell $c$ has a state $s_\psi$ defined by a set of semantic labels:
\begin{equation}
    s_\psi^{c} = \left[\ell_{Occ}, \ell_{Reli}, \ell_{Dyn}, \ell_{FoV}, \ell_{Sen}, \ell_{Occl}
    \right]
\end{equation}
\noindent being:
\begin{equation}
    \begin{array}{l}
        \ell_{Occ} = \{\text{\textit{occupied, free, unknown}}\} \\
        \ell_{Reli} = \{\text{\textit{reliable, unreliable}}\} \\
        \ell_{Dyn} = \{\text{\textit{static, oncoming, receding}}\} \\
        \ell_{FoV} = \{\text{\textit{in-view, M-FoV, O-FoV, F-FoV}}\} \\
        \ell_{Sen} = \{\text{\textit{sensed, unsensed}}\} \\
        \ell_{Occl} = \{\text{\textit{non-occluded, static, dynamic, unreliable}}\}
    \end{array}
\end{equation}
\noindent being M-FoV, O-FoV and F-FoV the maximum, occupied and free FoVs respectively.

The motivation behind this categorization and the necessary steps to compute the CG are presented in the following sections.

\subsection{Occupancy Categorization}
\label{sec:occupancy_division}

At the heart of the proposed space categorization lies a focus on occupancy state, aligning closely with the primary objective of OGs. Therefore, cells are divided by occupancy as follows:

\begin{equation}
    \label{eq:division_ocupacion}
    \ell_{Occ}^c = \left \{
    \begin{array}{ll}
        \text{\textit{occupied}}, & \text{if } m(O^c) \geq \mathpzc{T}_O \\  
        \text{\textit{free}}, & \text{if } m(F^c) \geq \mathpzc{T}_F \ \land \ m(O^c) < \mathpzc{T}_O \\
        \text{\textit{unknown}}, & \text{otherwise}
    \end{array}
    \right .
\end{equation}
\noindent where $\mathpzc{T}_O$ and $\mathpzc{T}_F$ are threshold values. 


\subsection{Occupied Space Categorization}
\label{sec:occupied_space_categorization}

DOGs already add a new layer of information to the occupied space (the dynamic state). Thus, its categorization seeks to effectively convey both estimations. With this intention, first, occupied cells are clustered into potential obstacles, then, these are evaluated concerning their dynamic state and reliability.


\subsubsection{Clustering}
\label{sec:clustering}

The categorization of \textit{occupied} space can be achieved by individually analyzing the data for each cell. However, DOGs are not free from noise and cells are usually assumed to be independent of each other---two cells corresponding to the same obstacles may exhibit different dynamic states \cite{2023_victor_framework, 2017_Piewak}. 
To filter noise and homogenize the data of adjacent cells, \textit{occupied} cells are clustered and their data is analyzed as a whole.

\textit{Occupied} cells are clustered according to proximity and velocity similarity. To achieve this, a Connected Components algorithm \cite{connected_components} is modified to include the evaluation of velocity differences. Cells within an \mbox{8-connected} neighborhood and a velocity difference smaller than $\mathcal{T}_v^{cl}$ are assumed to correspond to the same obstacle.

Each cluster $\varsigma$ obtained thhrough this process is described by the following state:
\begin{equation}
    \label{eq:categorization_cluster_state}
    \pmb{s}^\varsigma  = 
    \left [x_\text{x}, \ 
    x_\text{y}, \ 
    v, \ 
    \theta, \ 
    h, \
    n_t^\rho, \
    n_{obs}^{c}, \
    n^{c}
    \right]
\end{equation}
\noindent being $[x_\text{x}, x_\text{y}]$ the gravity center, $v$ and $\theta$ the cells' average velocity expressed in terms of module and orientation, $h$ the difference between the highest and lowest cells' height values, $n_t^\rho$ the average number of times that the particles gathered inside the cluster have been resampled, $n^{c}$ the number of cells of the cluster and $n_{obs}^{c}$ the number of cells of the cluster observed in the current iteration.

Clusters with less than $\mathpzc{T}^{cl}_{n^c}$ cells are considered noise. Thus, they are filtered and the category of their corresponding cells is changed from \textit{occupied} to \textit{unknown}.

\subsubsection{Occupied Space Dynamic Behavior} 
Depending on their dynamic behavior, obstacles can have varying degrees of relevance for the ego-vehicle. Indeed, as introduced earlier, many works classify obstacles into static and dynamic categories due to the heightened complexity of interactions with dynamic obstacles.

The proposed CG also uses the \textit{static--dynamic} differentiation but includes a further distinction regarding dynamic obstacles' direction in relation to the ego-vehicle: \textit{receding--oncoming}.

\textit{Occupied} cells receive the corresponding dynamic behavior label $\ell_{Dyn}$ based on their cluster's velocity and orientation. Therefore, for each cell $c$ belonging to the cluster $\varsigma$:
\begin{equation}    
    \label{eq:dynamic_evaluation}
    \ell_{Dyn}^{c, \varsigma} = \left \{
    \begin{array}{ll}        
        \text{\textit{static}}, & \text{if } v^\varsigma < \mathpzc{T}_v^{static} 
        \\        
        \text{\textit{oncoming}}, & \text{if } v^\varsigma \geq \mathpzc{T}_v^{static} \land \theta^{\varsigma\rightarrow ego} \leq \mathpzc{T}_\theta^{onc}
        \\
        \text{\textit{receding}}, & \text{if } v^\varsigma \geq \mathpzc{T}_v^{static} \land \theta^{\varsigma\rightarrow ego} > \mathpzc{T}_\theta^{onc}
    \end{array}
    \right .
\end{equation}
\noindent where: 
\begin{equation}
    \theta^{\varsigma\rightarrow ego} = \left|\theta^\varsigma - \text{atan}\left(\frac{x_\text{y}^{ego} - x_\text{y}^\varsigma}{x_\text{x}^{ego} - x_\text{x}^\varsigma}\right)\right| 
\end{equation} 
\noindent being $[x_\text{x}^{ego}, x_\text{y}^{ego}]$ the position of the \mbox{ego-vehicle}, $\mathpzc{T}_v^{static}$ a velocity threshold denoting static obstacles and \mbox{$\mathpzc{T}_\theta^{onc} < 90^\circ$} an angular threshold denoting if the obstacle is heading towards the ego-vehicle. 

\subsubsection{Occupied Space Reliability} 

Like any estimation algorithm, DOGs are susceptible to noise (e.g. incorrect point cloud obstacle--ground segmentation). Moreover, they require a certain convergence time to provide reliable estimations.

Seeking to denote the \textit{occupied} cells whose estimation is not entirely reliable three heuristic conditions are evaluated: 
\begin{itemize}
    \item Height: One of the strongest criteria for obstacle detection in point clouds is the presence of vertical structures \cite{2021_Victor_suelo}. Therefore, a minimum height value $\mathpzc{T}_{height}^{reli}$ is set to filter possible incorrect obstacle-ground classification.

    \item Currently observed: Cells with occupied mass, yet currently unobserved by the sensors, may correspond to erroneous obstacle hypotheses or to inaccurate position and velocity estimations. Therefore, a minimum percentage of observed cells $\mathpzc{T}_{obs}^{reli}$ supporting the obstacle hypothesis is defined. 

    \item Particles' age: The convergence time required to obtain reliable estimations is taken into account by setting a minimum age $\mathpzc{T}_{age}^{reli}$, approximated as the number of times that the particles have been resampled.
\end{itemize}
\begin{equation}
    \label{eq:reliability}
     \left(h^\varsigma < \mathpzc{T}_{height}^{reli}\right) \lor \left(\frac{n_{obs}^{c,\varsigma}}{n^{c,\varsigma}} < \mathpzc{T}_{obs}^{reli}\right) \lor \left(n_t^{\rho,\varsigma} < \mathpzc{T}_{age}^{reli}\right) 
\end{equation}

\textit{Occupied} cells corresponding to a cluster that does not fulfill the above-mentioned conditions (\ref{eq:reliability}) receive the label $\ell_{Reli}^{c, \varsigma} = \text{\textit{unreliable}}$. The rest are labeled opposingly as \textit{reliable}.

\begin{figure}[!htbp]
    \centering
    \includegraphics[width=0.85\columnwidth]{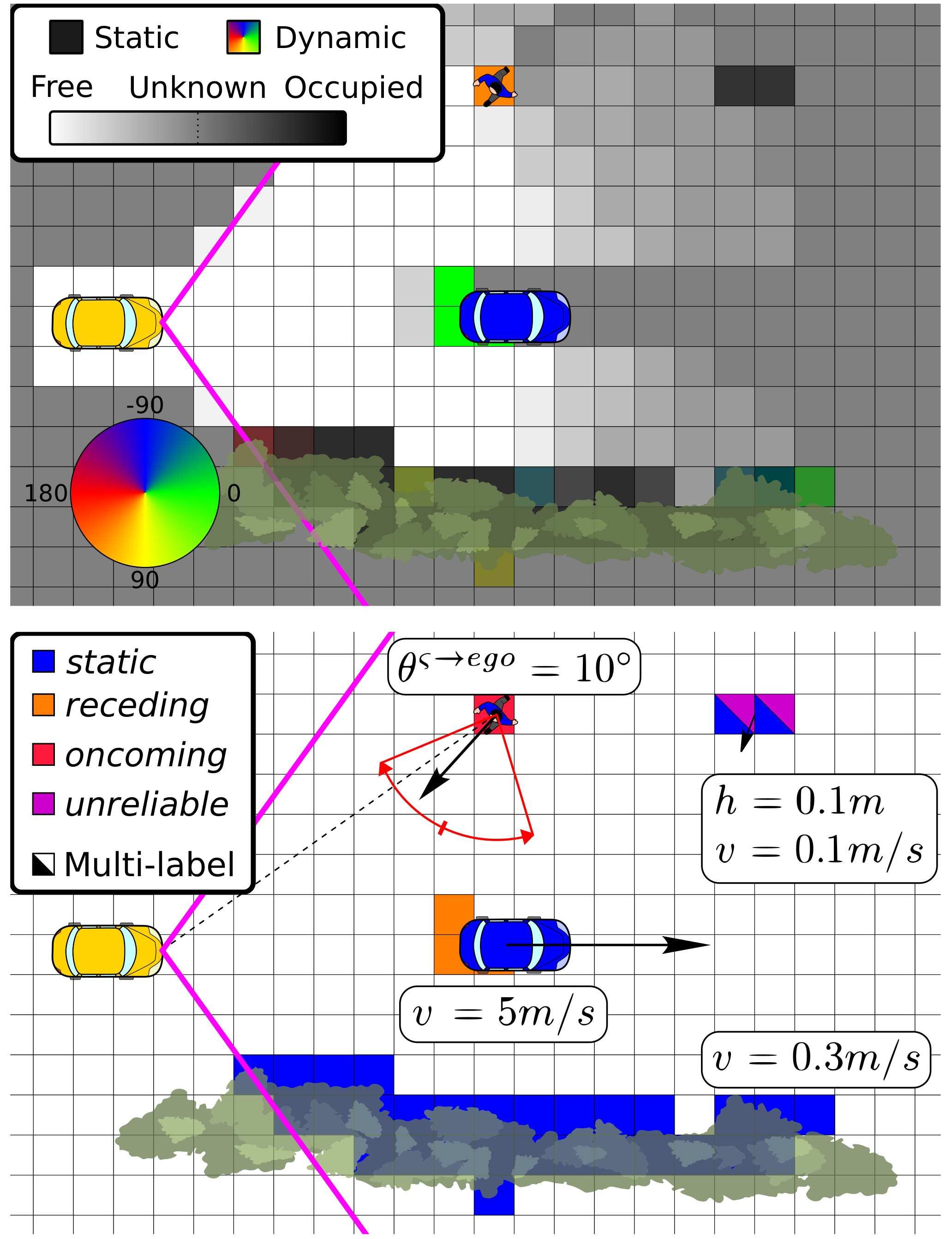}    
    \caption{Illustrative example of \textit{occupied} space categorization. 
    For better visualization, \textit{reliable} cells are only drawn regarding their dynamic behavior. 
    }
    \label{fig:ejemplo_categorizacion_ocupado}
\end{figure}

Figure~\ref{fig:ejemplo_categorizacion_ocupado} shows an illustrative example that summarizes the categorization of the occupied space.

\subsection{Unknown Space Categorization}
\label{sec:unknown_space_categorization}

In contrast to the occupied space and as explained in Section~\ref{sec:introduction}, OGs generally do not provide further data concerning the unknown space. This work seeks to add new information about it by analyzing the different factors that may constrain the ability of the perception system to perceive the environment.

With this intention, the evaluation of unknown space involves (i)~the perception system's FoV, (ii)~the actually sensed space, and (iii)~the occlusions caused by obstacles in the scene.

\subsubsection{Perception System FoV} 

It is commonly assumed that the FoV of a perception system is equal to the union of its sensors' FoVs. Nevertheless, every perception system has limitations that cause the modeling of unknown space inside this area even under ideal performance. For instance, the time needed for convergence, the maximal reliability of the observations, or design parameters such as the minimum acceptable confidence in the estimation---e.g. $\mathpzc{T}_O$ and $\mathpzc{T}_F$ in (\ref{eq:division_ocupacion}). 
Therefore, in this work, it is considered that the union of the sensors' FoV defines the ``maximum FoV'' (M-FoV), while the aforementioned space delimited by the constraints denotes the area that the perception system can ideally estimate with confidence (``occupancy FoV'').

In OGs for autonomous driving applications, there is interest in both occupied space and free space. However, different constraints are typically defined for them---usually seeking to avoid false negatives for occupied space while being more cautious with free space. For example, \cite{2023_Victor_comparison} sets different maximum values for occupied and free measurements, \cite{2013_Nuss} defines a valid maximum height range for free space, and \cite{2007_Weiss} models the occupied space based on the number of impacts in a cell and weights the free space based on the distance to the sensor. 
Therefore, in addition to the M-FoV, this work also considers a FoV for occupied space estimation (O-FoV) and a FoV for free space estimation (F-FoV).

\begin{figure}[htbp]
    \centering
    \includegraphics[width=\columnwidth]{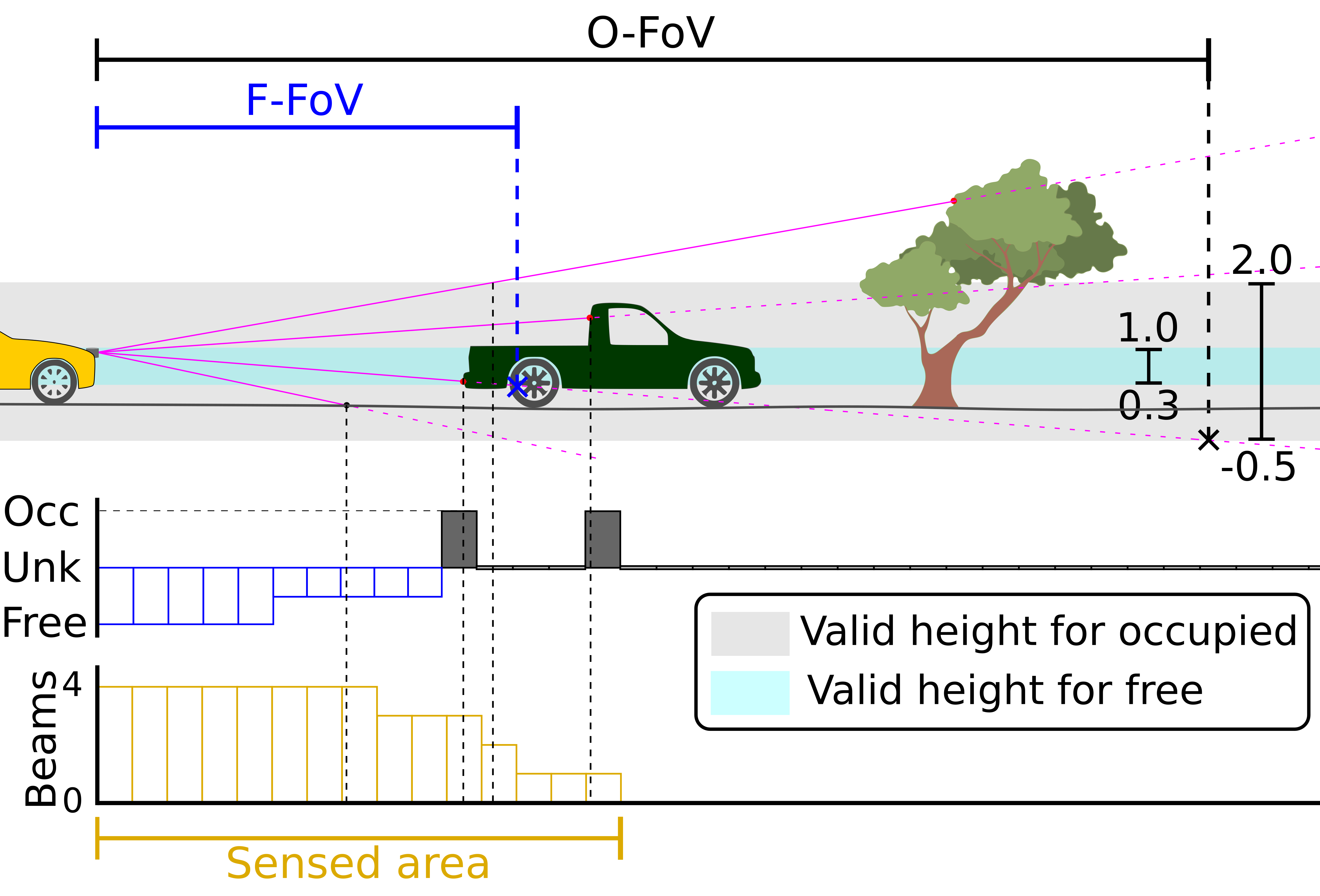}    
    \caption{Illustrative example of the O-FoV, F-FoV and Sensed area calculation. The perception system used has a four-layer LiDAR sensor and two valid height ranges for occupied and free space measurements. 
    O-FoV and F-FoV are calculated for $n_{iter} = 1$, so they only take into account the occupancy observation, which is primarily determined by the height ranges, i.e. they cover the space until the laser beams surpass them. 
    The Sensed area is computed as the set of cells traversed by beams with valid information, i.e. it is limited by the obstacles, the ground and the maximum height value.    
    }
    \label{fig:FoV}
\end{figure}

The set of cells $\pmb{C}_{\text{\textit{M-FoV}}}$ within the M-FoV is computed by directly evaluating the position of each cell with respect to the sensors' FoV. In contrast, the O-FoV and F-FoV are approximated by evaluating the outcome of the perception system under ideal conditions with simulation tests in which all the space is observed as occupied or free, respectively. $\pmb{C}_{\text{\textit{O-FoV}}}$ and $\pmb{C}_{\text{\textit{F-FoV}}}$ are defined as the set of cells that after $n_{iter}$ iterations fulfill the occupancy categorization (\ref{eq:division_ocupacion}). 
Note that these areas are computed based on an elapsed time ($n_{iter}$). The number of iterations can be fixed (e.g. just one iteration to obtain the FoVs relative to the observation or the maximum assumable time without estimating the environment) or dynamically adjusted (e.g. when stopped at an intersection, it can be increased to evaluate how far the estimation can reach).
Figure~\ref{fig:FoV} shows an illustrative example of the O-FoV and F-FoV calculation. 

By definition, \mbox{O-FoV} and \mbox{F-FoV} should always be contained within \mbox{M-FoV}. Furthermore, generally, \mbox{F-FoV} is smaller than \mbox{O-FoV} for the reasons explained above. Therefore, in order to categorize \textit{unknown} cells due to the perception system's FoVs, a priority order is defined:
\begin{equation}
{\small
    \ell_{FoV}^c = \left \{
        \begin{array}{ll}            
         \text{\textit{M-FoV}} & \text{if } c \notin \pmb{C}_{\text{\textit{M-FoV}}} 
         \\
         \text{\textit{O-FoV}} & \text{if } c \in \pmb{C}_{\text{\textit{M-FoV}}} \land c \notin \pmb{C}_{\text{\textit{O-FoV}}} 
         \\
         \text{\textit{F-FoV}} & \text{if } c \in \pmb{C}_{\text{\textit{M-FoV}}} \land c \in \pmb{C}_{\text{\textit{O-FoV}}} \land c \notin \pmb{C}_{\text{\textit{F-FoV}}} 
         \\
         \text{\textit{in-view}} & \text{otherwise}
        \end{array}
    \right .     
}    
\end{equation}
Notice that \textit{unknown} cells inside the three FoVs cannot be explained by means of the perception system's FoV. Therefore, they are labeled as \textit{in-view}.

\subsubsection{Sensed Area}
\label{sec:sensed_area}

Real-world autonomous driving scenes are not ideal. Typically, there are several factors that affect the perception system by diminishing the expected perceived area, e.g. obstacles, variable terrain, weather conditions, etc. 

To denote the space that is actually being perceived, the Sensed area is computed in each iteration as the set of cells traversed by the laser beams with valid information. This area can be calculated either by raytracing each laser beam or by directly leveraging the calculation process of the DOG's occupancy observation step. 
Notice that the modeled Sensed area must be compliant with the observation method used, e.g. height limits or the shape assumed for the laser beams.
Figure~\ref{fig:FoV} compares the calculation of Sensed area with the expected O-FoV and F-FoV.  

\begin{figure}[!htbp]
    \centering
    \includegraphics[width=1\columnwidth]{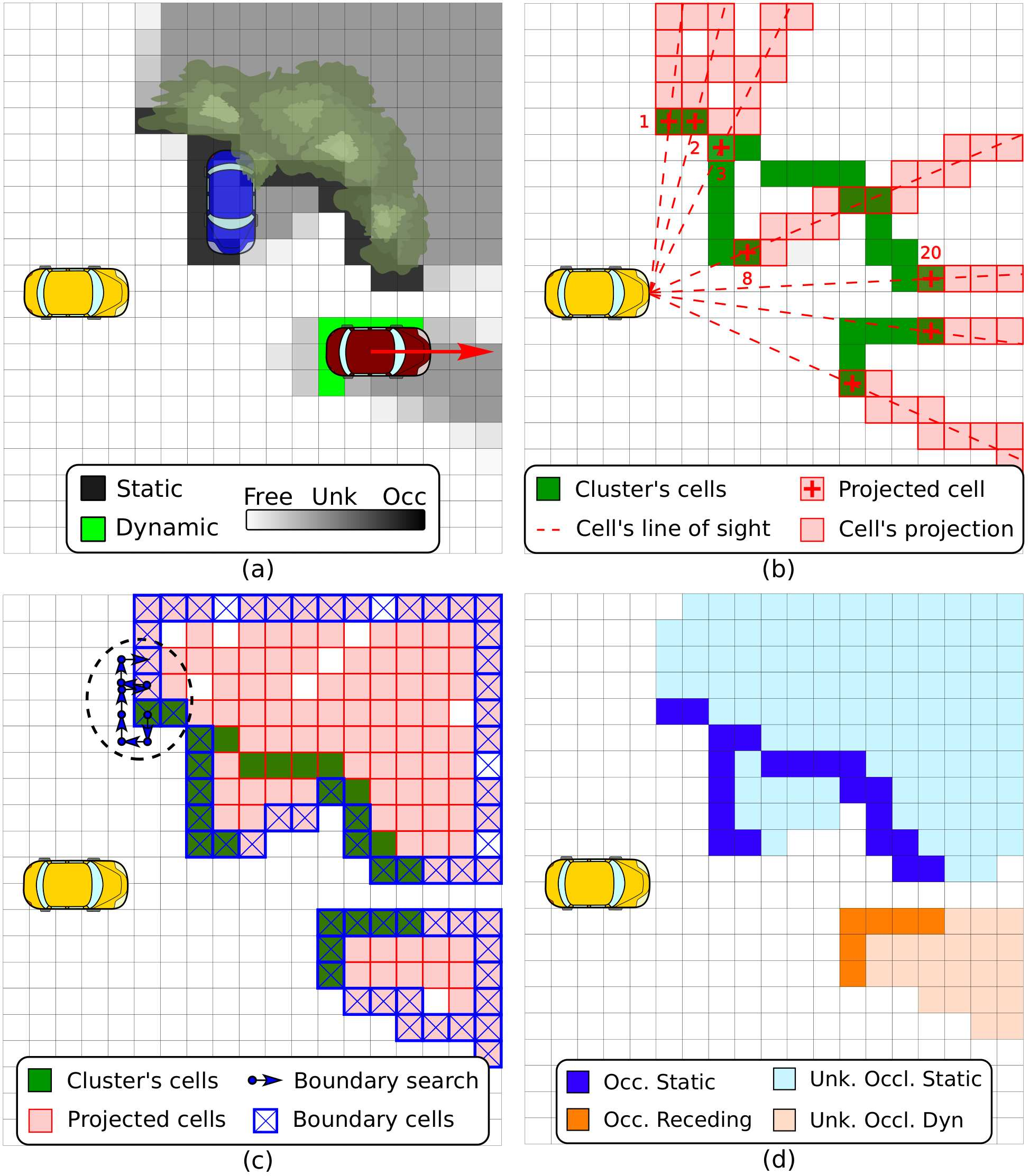}    
    \caption{Illustrative example of the process for computing the occlusion produced by a cluster with a complex form. 
    a) Estimated occupancy. b) Example of some cluster's cells projected by their line of sight. c) Occluded space boundary calculation using the traversed cells and grid borders. d) Cell categorization considering occlusions' boundaries and obstacles' dynamic state.}
    \label{fig:example_oclusion}
\end{figure}

The information of the Sensed area is added to the CG through the label $\ell_{Sen}$. 
\textit{Unknown} cells outside the Sensed area may be due to this lack of data and thus labeled accordingly as \textit{unsensed}. 
The remaining cells are set as \textit{sensed}.

\subsubsection{Occlusions}
\label{sec:occlusions}

The presence of obstacles in the scene diminishes the space perceived by the sensors by partially or completely occluding the space behind them. 
The Sensed area is capable of identifying areas where no laser beams traverse, but it does not attribute this absence of beams to its causes.
Moreover, by associating the occluded space with its corresponding obstacle, it can be inferred whether the occlusion is temporary (caused by a dynamic obstacle) or permanent (caused by a static obstacle).

Therefore, the corresponding occluded areas are computed for each cluster $\varsigma$. 
First, Bresenham's algorithm \cite{Bresenham} is used to project every border cell of the cluster along its line of sight. 
Then, the boundary of all cells traversed in this way is computed using the Moore-Neighbor tracing algorithm \cite{Moore}. The set of cells $\pmb{C}_{occl}^\varsigma$ within this boundary corresponds to the space occluded by the cluster. Fig.~\ref{fig:example_oclusion} shows an illustrative example of this occlusion calculation process.

As mentioned previously, occlusions can be considered temporary or permanent depending on the dynamic state of the obstacles. For example, the space occluded by a dynamic vehicle changes as it moves, while the space occluded by a building will remain unknown unless the ego-vehicle itself moves. Consequently, \textit{unknown} cells within the set of occluded cells $\pmb{C}_{occl}^\varsigma$ are labeled according to the velocity and reliability of the corresponding cluster $\varsigma$:
\begin{itemize}
    \item In the case the cluster is \textit{unreliable} (\ref{eq:reliability}), no trustworthy dynamic information can be assigned to the occlusion, i.e. cells are labeled as $\ell_{occl}^{c, \varsigma} = \text{\textit{occl. unreliable}}$.

    \item In the case the cluster is \textit{reliable}, the occlusion is labeled according to its dynamic state: 
    \begin{equation}
        \ell_{occl}^{c, \varsigma} = \left \{
            \begin{array}{ll}
                \text{\textit{occl. static}} & \text{if } v^\varsigma < \mathpzc{T}_v^{static}\\
                \text{\textit{occl. dynamic}} & \text{otherwise} 
            \end{array}
        \right .
    \end{equation}
\end{itemize}

Cells can be occluded by more than one obstacle. In these cases, the predominant label follows a temporal and reliability criteria: 1st \textit{occl. static}, 2nd \textit{occl. dynamic}, 3rd \textit{occl. unreliable}.

\begin{figure}[!htbp]
    \centering
    \includegraphics[width=0.9\columnwidth]{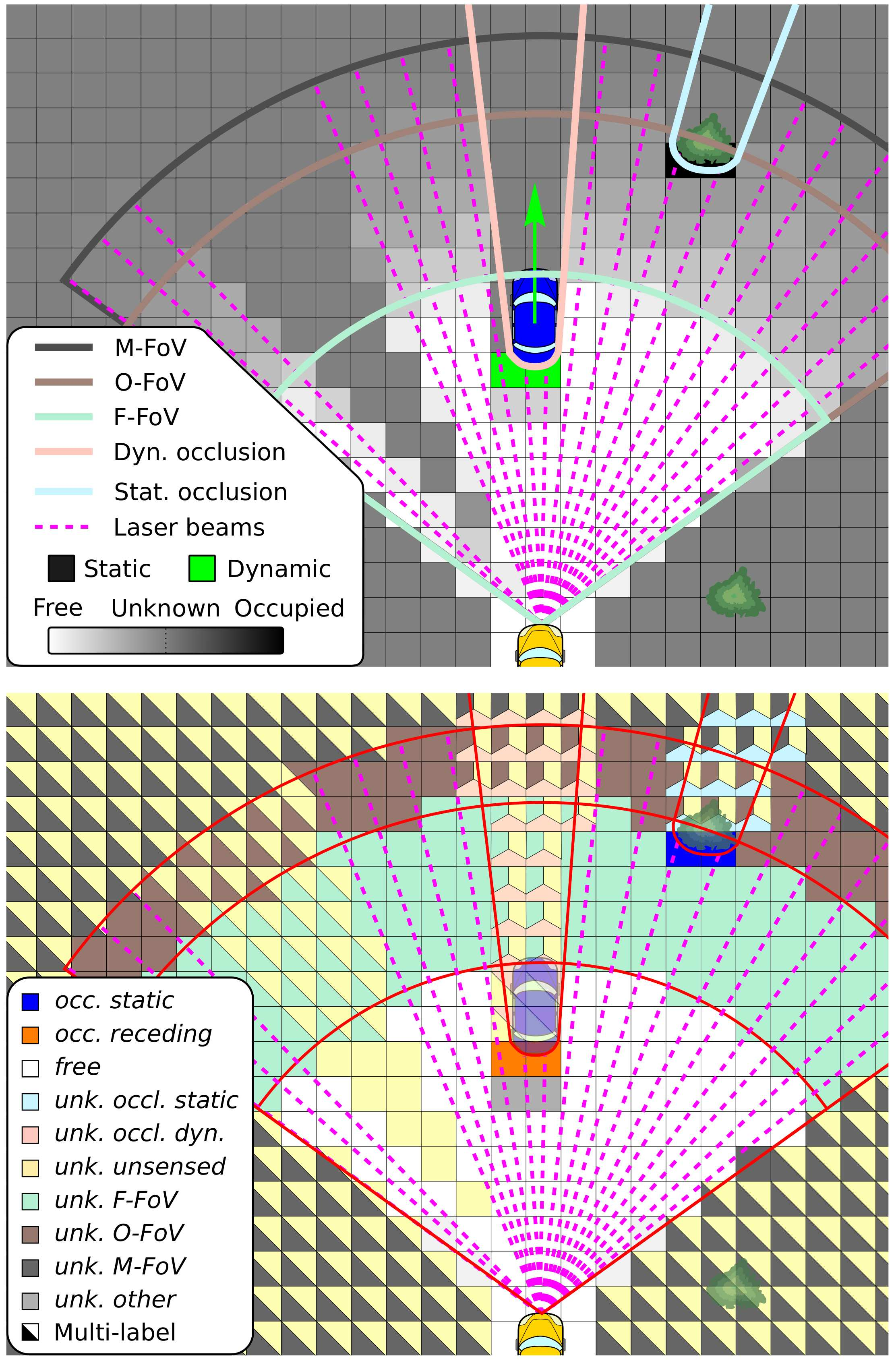}
    \caption{CG calculation example with special emphasis on the unknown space categorization.   
    \textit{Unknown} cells are only visualized regarding their possible causes, i.e. labels \textit{visible}, \textit{sensed} and \textit{non-occluded} are not represented. In the case of \textit{unknown} cell with these three labels at the same time, it is considered that the causes are unidentified and an auxiliary label \textit{other} is set.}  \label{fig:ejemplo_motivacion_categorizacion_unknown}
\end{figure}

Fig.~\ref{fig:ejemplo_motivacion_categorizacion_unknown} shows an illustrative example of the calculation of the CG with special emphasis on the categorization of the unknown space.

\section{Experimental Results}
\label{sec:experiments}

The work presented in this paper has been tested with an autonomous vehicle prototype of the AUTOPIA group \cite{2015_AUTOPIA} equipped with a set of sensors that allow to estimate its own state and the surrounding environment. 
Details about the perception system used can be found in \cite{2023_victor_framework, 2023_tesis_Victor}.

Seeking to provide a clear visualization of the CG calculation with a particular focus on categorizing the unknown space, the environment is estimated using only a 4-layer Ibeo-LUX LiDAR sensor \cite{ibeo_2010}. 
This sensor has the field of view and layer arrangement specified in Tab~\ref{tab:ibeo}, which is very similar to those shown previously in Figs.~\ref{fig:ejemplo_categorizacion_ocupado}, \ref{fig:FoV} and \ref{fig:ejemplo_motivacion_categorizacion_unknown}.

In the following, three illustrative experiments are analyzed. Further results are provided in a video available in \url{https://youtu.be/qTdJmH1nuLk}.

\begin{table}[!htbp]
\centering
\caption{Specifications of Ibeo-LUX LiDAR sensor setup for the calculation of the FoVs and Sensed area.}
\label{tab:ibeo}
\begin{tabular}{|l|c|}
\hline
\textbf{Nº of layers}                    & 4                                          \\ \hline
\textbf{Vertical FoV}                    & $[-1.2,\ -0.4,\ 0.4,\ 1.2]^\circ = 3.2^\circ$ \\ \hline
\multirow{2}{*}{\textbf{Horizontal FoV}} & \ 2 layers: $[-50,\ 50]^\circ = 100^\circ$     \\ \cline{2-2} 
                                         & 2 layers: $[-50,\ 35]^\circ = 85^\circ$     \\ \hline
\textbf{Angular resolution}              & $0.25^\circ$                              \\ \hline
\end{tabular}
\end{table}

\begin{figure*}[htbp]
    \centering
    \includegraphics[width=\textwidth]{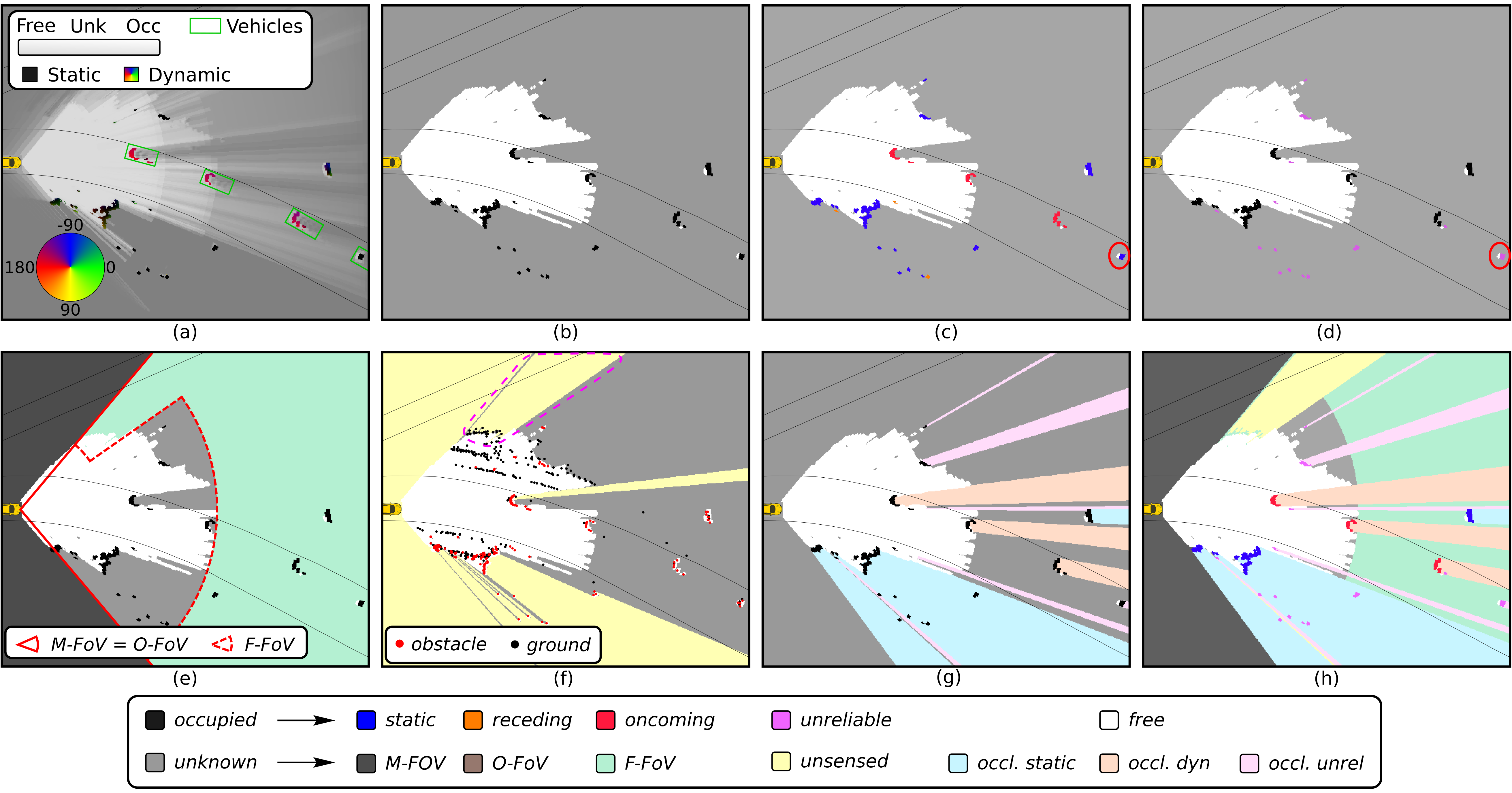}    
    \caption{Real-scenario CG calculation example. Subfigures b) to g) show the six different labels assigned to the cells in order to categorize the space. 
    (a)~DOG and manually labeled vehicles. 
    (b)~Occupancy division. 
    (c)~Occupied space based on dynamic behavior. 
    (d)~Occupied space based on reliability. 
    (e)~Unknown space based on perception system's FoV.
    (f)~Unknown space based on Sensed area and LiDAR point cloud classified as obstacle or ground.
    (g)~Unknown space based on occlusions.
    (h)~CG depicted with the most relevant label for each cell.}
    \label{fig:1725}
\end{figure*}

The first experiment, Fig.~\ref{fig:1725}, illustrates the process of computing the CG by breaking it down into all possible labels. The scene consists of the ego-vehicle traveling along a two-way urban road surrounded by trees at the right and grass at the left. There are no other vehicles in the ego lane, but four vehicles are approaching through the adjacent lane.
Fig.~\ref{fig:1725}a shows the initial DOG estimation.

Fig.~\ref{fig:1725}b provides the first categorization step: occupancy-based division. It can be clearly seen that the perception system used has limited capabilities in estimating the environment. Although it is able to estimate the four vehicles and the free space ahead, it also models large areas of unknown space.

The further categorization of the occupied space is visualized in Fig.~\ref{fig:1725}c (dynamic behavior) and Fig.~\ref{fig:1725}d (estimation reliability). 
As can be noticed, the three closest vehicles have been correctly estimated and labeled as \textit{occupied}, \textit{oncoming} and \textit{reliable}. In contrast, the farthest vehicle (highlighted with a red circle) has just entered the DOG's range, resulting in an erroneous static state, i.e. \textit{occupied} and \textit{static}. However, the young age of its corresponding occupied cells is noticed during the reliability evaluation defined in eq.~(\ref{eq:reliability}), leading also to the label \textit{unreliable}.
With regard to the vegetation on both sides of the road, it is susceptible to noise due to its variable patterns and possible gaps through which laser beams may pass through. Thus, while most are correctly labeled as \textit{occupied} and \textit{static}, the reliability conditions are triggered, indicating that various groups of cells are also deemed \textit{unreliable}.

In Fig.~\ref{fig:1725}e the categorization of the \textit{unknown} cells concerning the FoVs of the perception system is shown. In the DOG implementation used, LiDAR points classified as obstacles are considered reliable measures of occupied space. Therefore, M-FoV and O-FoV cover the same area, i.e. they are both delimited by the sensors' angular FoV and DOG's size. In contrast, the F-FoV is much smaller, as a conservative convergence time of $n_{iter} = 2$ iterations is defined and the DOG employed has a free space observation model based also on conservative maximum and minimum height values (similar to Fig.~\ref{fig:FoV}, see \cite{2023_tesis_Victor} for more details). 
Thus, at farther distances, only one layer can pass over the ground while still adhering to the maximum height criterion, which is insufficient to satisfy the minimum confidence value $\mathpzc{T_F}$ for the free space.

In contrast to the FoVs, the current Sensed area, Fig.~\ref{fig:1725}f, denotes how the ideal perception of the environment is limited by the different elements of the scene (obstacles, variable terrain, etc.). In most instances, this limitation arises from obstacle occlusions, which can be identified and further categorized based on the corresponding obstacle's state (\textit{static}, \textit{dynamic} or \textit{unreliable})---Fig.~\ref{fig:1725}g. However, there is also an area of grass (highlighted in magenta) where the laser beams hit the irregular terrain, preventing the capture of information beyond it.  

Lastly, to provide a comprehensive overview of the computed CG, all categorizations are displayed together in Fig.~\ref{fig:1725}h. However, only one label is selected for each cell based on the following relevance criteria:
\begin{itemize}
    \item \textbf{\textit{Occupied} cells:} If the cell is \textit{unreliable} it is drawn accordingly since its occupancy and dynamic states may be erroneous. Otherwise, the label corresponding to the dynamic behavior is used.

    \item \textbf{\textit{Unknown} cells:} Only the labels denoting the possible causes are visualized. Occlusion-based labels take precedence over the rest of labels, as they provide additional information about the obstacles causing them. Following them, the \textit{M-FoV} label is prioritized, as it defines the maximum area where laser measurements are feasible. Then, the \textit{unsensed} label predominates over the \textit{O-FoV} and \textit{F-FoV} labels, as it indicates the current perceived area and not the theoretical sensing space.    
    Lastly, in the case of \textit{unknown} cells labeled as \textit{visible}, \textit{sensed} and \textit{not-occluded} simultaneously, it is considered that the causes are unidentified, and the auxiliary label \textit{other} is assigned.
\end{itemize}

As can be noticed, the proposed CG is able to properly convey DOG's occupancy and dynamic estimations, while also identifying partially trustworthy occupied cells. 

Additionally, the unknown space categorization allows for better situation understanding from the point of view of the perception system capabilities. 
On the one hand, the FoVs-based analysis indicates that the perception on both sides of the ego-vehicle will always be limited, while at the front, at long distances, although obstacles can be detected, ensuring their absence (free space) is more challenging. On the other hand, the analyses of the Sensed area and the occlusions reveal that the visibility concerning the road is almost unaffected by environmental factors---
i.e., most of the unknown space within it is attributed to the limitations in estimating free space (F-FoV).

\begin{figure}[!htbp]
    \centering
    \includegraphics[width=\columnwidth]{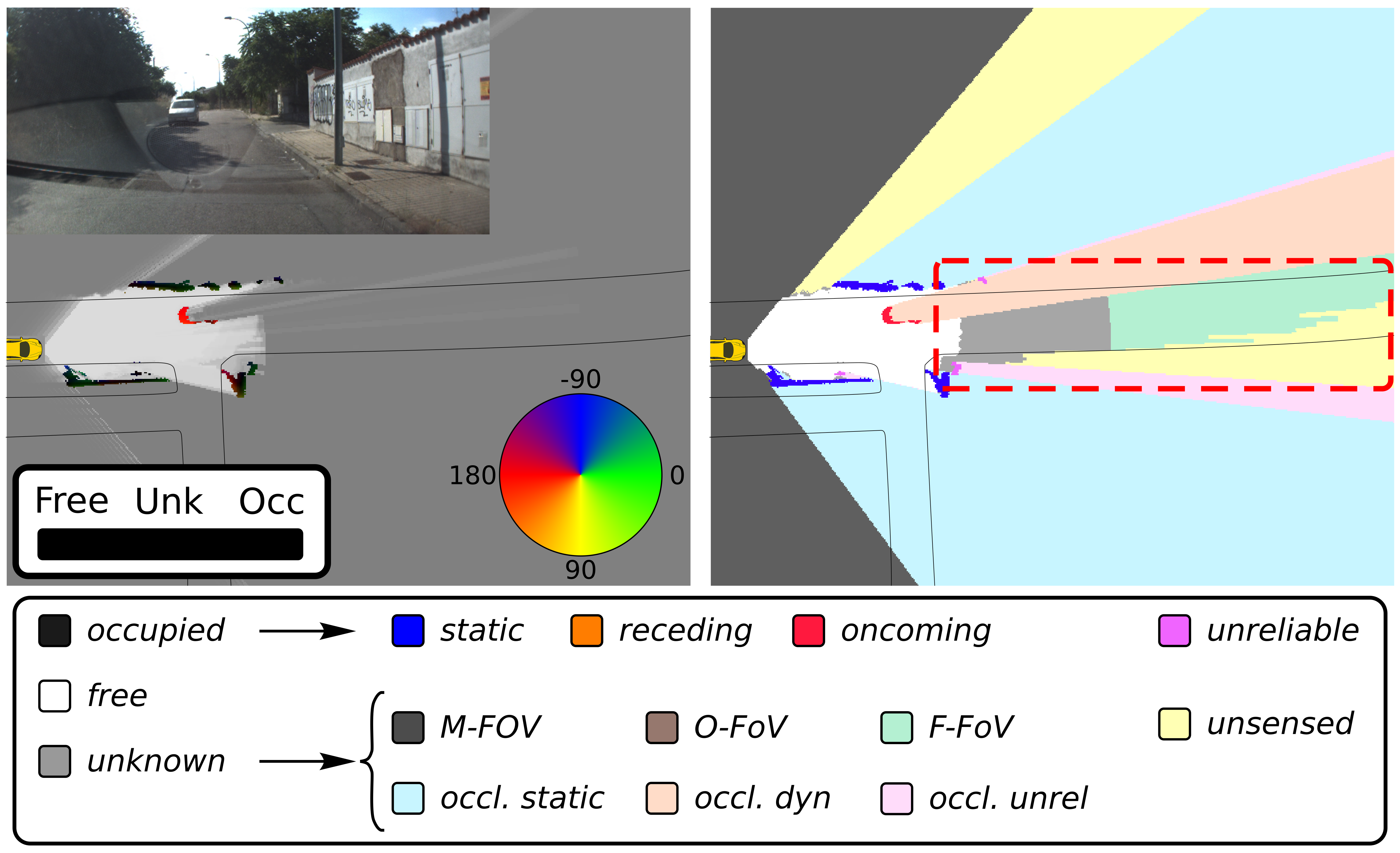}    
    \caption{CG computed for a scenario with a significant amount of unexpected unknown space. Only the most relevant label for each cell is visualized.}
    \label{fig:ibeo_2865}
\end{figure}

Fig.~\ref{fig:ibeo_2865} shows another scene particularly useful to illustrate the utility of the occlusion-based categorization. On the one hand, the space occluded by the oncoming vehicle is labeled as \textit{occl. dynamic} indicating a temporary occlusion that changes as this vehicle moves and eventually disappears.
On the other hand, the road exit located in front and to the right of the ego-vehicle is completely occluded by a static obstacle (\textit{occl. static}). In this case, this occlusion is permanent, requiring the ego-vehicle to move in order to perceive the corresponding area. 

Additionally, this experiment also allows to expose how the CG may be used to identify challenging situations from the perspective of the perception system. A significant unknown area is modeled within the road ahead of the ego-vehicle (highligted with a red dashed line). Part of this area can be attributed to the free space FoV (\textit{F-FoV}) aligning with expectations. However, at a closer distance, the causes of the unknown space remain unidentified (\textit{other})---it lies within the three FoVs, without occlusions, and laser beams are sensing the space. Moreover, the farthest part falls within the M-FoV, also without occlusions, yet the LiDAR is unable to capture any information (\textit{unsensed}). Consequently, it can be inferred that something unexpected is diminishing the perception system estimation capabilities. The real cause of this unknown space is the sloped terrain; laser beams hit the ground earlier, thus limiting the number of layers that gather information at longer distances.

\begin{figure}[htbp]
    \centering
    \includegraphics[width=\columnwidth]{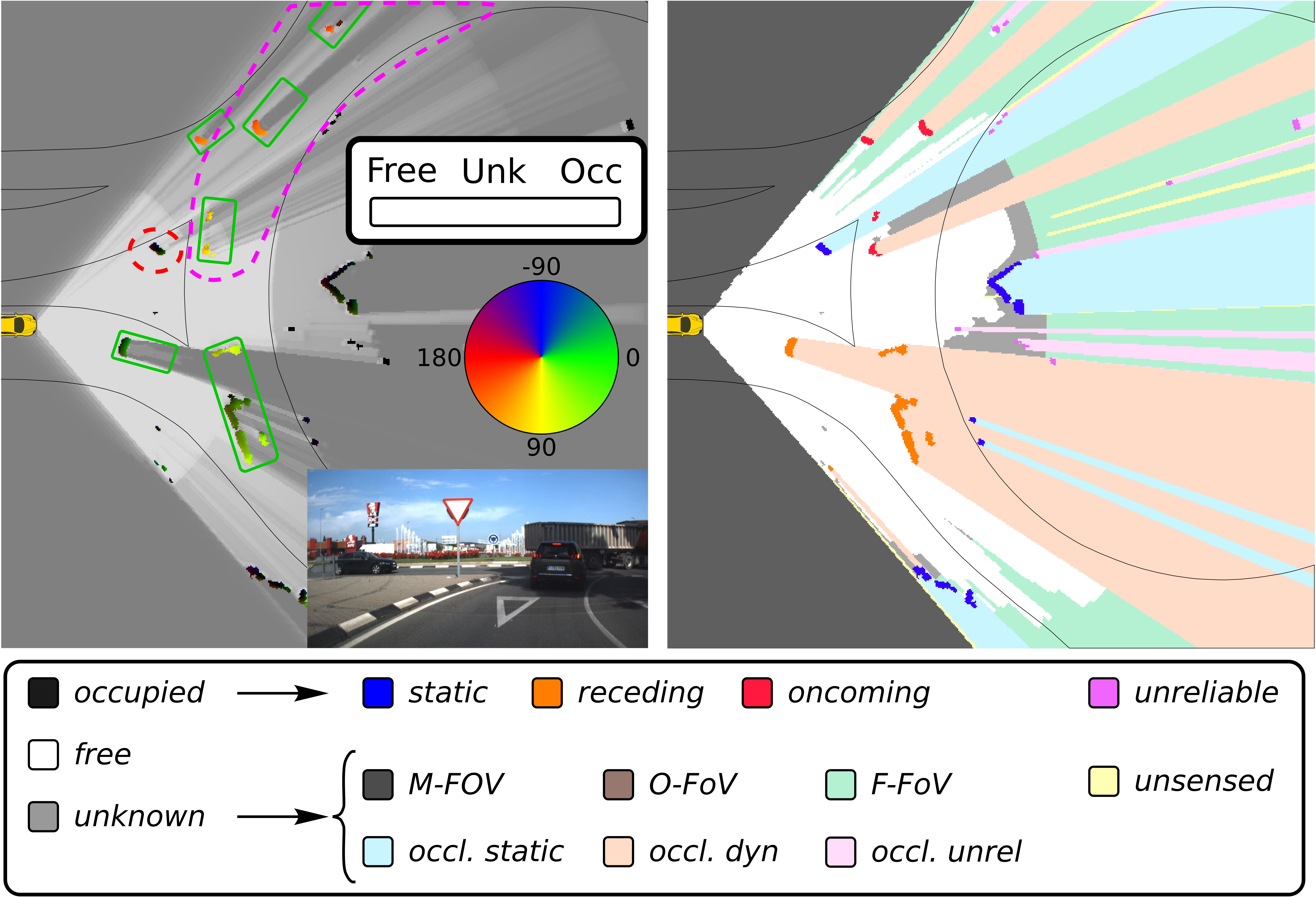}    
    \caption{CG computed for a dense traffic scene. Only the most relevant label for each cell is visualized.}
    \label{fig:ibeo_3562}
\end{figure}

Lastly, Fig.~\ref{fig:ibeo_3562} shows a scene in which the ego-vehicle has to enter a roundabout with dense traffic. The area of interest to evaluate the incorporation is highlighted with a magenta dashed line. The F-FoV indicates that estimating free space in this area under the desired convergence time is difficult even under ideal conditions. Moreover, several obstacles occlude the space. 
While most of these occlusions are caused by dynamic obstacles, i.e. temporary, there is a traffic signal (marked with a red circle) that leads to a static occlusion, which is permanent. 
Thus, this scene can also be considered challenging from the perspective of the perception system, as there are elements in the scene and inherent limitations of the perception system that affect the estimation of the main target area.

\section{Conclusions}
\label{sec:conclusions}

In this work, a Categorized Grid has been proposed to convey the estimation of a DOG and provide additional insights about the modeled unknown space. 

First, cells are divided based on their occupancy state. Then, \textit{occupied} cells are further categorized according to their dynamic state and estimation reliability, and \textit{unknown} cells with respect to their possible causes.
This categorization seeks to provide a better situation understanding from the perspective of the perception system. On one hand, labeling the cells based on their features facilitates the comprehension and management of DOG's information. On the other hand, the proposed unknown space analysis provides insights into the potential causes influencing its estimation, encompassing design constraints, temporary or permanent elements of the scenario, or unidentified factors. 

Future works will explore the filtering over time for the labels assigned to the cells in order to obtain a robust categorization.

\addtolength{\textheight}{-12cm}   


\bibliographystyle{ieeetr}

\end{document}